\documentclass{article}
\usepackage{ijcai19}

\usepackage{times}
\usepackage{soul}
\usepackage{url}
\usepackage[hidelinks]{hyperref}
\usepackage[utf8]{inputenc}
\usepackage[small]{caption}
\usepackage{graphicx}
\usepackage{amsmath}
\usepackage{booktabs}
\usepackage{algorithm}
\usepackage{algorithmic}
\usepackage{enumerate}
\usepackage{array}
\usepackage{threeparttable}
\usepackage{multirow}
\usepackage{subfigure}
\title{Deep Global-Relative Networks for End-to-End 6-DoF Visual Localization and Odometry}

\author{
    Yimin Lin, Zhaoxiang Liu, Jianfeng Huang, Chaopeng Wang, Guoguang Du, Jinqiang Bai*,
Shiguo Lian, Bill Huang
    \affiliations
    AI Department, CloudMinds Technologies Co. Ltd, Beijing, China.
    *School of Electronic Information Engineering, Beihang University, Beijing, China\emails
    \{anson.lin, robin.liu, jianfeng.huang.interns, chaopeng.wang, george.du, scott.lian, bill\}@cloudminds.com.
    *baijinqiang@buaa.edu.cn
}

\begin{document}

\maketitle

\begin{abstract}
Although a wide variety of deep neural networks for robust Visual Odometry (VO) can be found in the literature, they are still unable to solve the drift problem in long-term robot navigation. Thus, this paper aims to propose novel deep end-to-end networks for long-term 6-DoF VO task. It mainly fuses relative and global networks based on Recurrent Convolutional Neural Networks (RCNNs) to improve the monocular localization accuracy. Indeed, the relative sub-networks are implemented to smooth the VO trajectory, while global sub-networks are designed to avoid drift problem. All the parameters are jointly optimized using Cross Transformation Constraints (CTC), which represents temporal geometric consistency of the consecutive frames, and Mean Square Error (MSE) between the predicted pose and ground truth. The experimental results on both indoor and outdoor datasets show that our method outperforms other state-of-the-art learning-based VO methods in terms of pose accuracy.
\end{abstract}

\section{Introduction}\label{sec:Intro}

The problem of visual localization has drawn significant attention from many researchers over the past few decades. Solutions for overcoming this problem come from computer vision and robotic communities by means of Structure from Motion (SfM) and visual Simultaneous Localization and Mapping (vSLAM)~\cite{1cadena2016past,2ozyecsil2017survey}. Many variants of these solutions have started to make an impact in a wide range of applications, including autonomous navigation and augmented reality.

During the past few years, most of traditional visual localization techniques have been proposed and grounded on the estimate of the camera motion among a set of consecutive frames with geometric methods. For example, the feature-based method uses the projective geometry relations between 3D feature points of the scene and their projection on the image plane~\cite{3ORBSLAM,4PTAM}, or the direct method minimizes the gradient of the pixel intensities across consecutive images~\cite{5engel2014lsd,6engel2018direct}. However, these techniques are critical to ideal and controlled environments, e.g., with a large amount of texture, unchanged illumination and without dynamic objects. Obviously, their performance drops quickly when facing those challenging and unpredicted scenarios.


Recently, a great breakthrough has been achieved in the Deep Learning (DL), through the application of Convolutional Neural Networks (CNNs) and Recurrent Neural Networks (RNNs), e.g., for the object recognition and scene classification tasks. Therefore, learning-based visual odometry in the past few years has seen an increasing attention of the computer vision and robotic communities ~\cite{23li2018ongoing}. This is due to its potentials in learning capability and the robustness to camera parameters and challenging environments. However, so far they are still unable to outperform most state-of-the-art feature-based localization methods. The drift from the true trajectory due to accumulation of errors over time is inevitable in those learning based VO system. This is due to the fact that such approaches cannot exploit high-capacity learning 3D structural constraints from limited training datasets. Recent work ~\cite{14clark2017vidloc} concluded that global place recognition and camera relocalization plays a significant role in reducing these global drifts. As demonstrated in another relevant VLocNet ~\cite{12valada2018deep}, the global and Siamese-type relative networks are designed for inferring global poses with the great help of relative motion. Nevertheless, VO drift problem still exists since its global and relative networks are separately optimized and regressed by a multitask alternating optimization strategy.

To solve the drift problem completely, this paper extends VLocNet to fuse both relative and global networks, and considers more temporal sequences with LSTM incorporated in each networks for accurate pose prediction. Furthermore, we also employ a geometric consistency of the adjacent frames for regressing the relative and global networks at the same time. This proposed method brings two advantages: one is obviously that we leverage the camera re-localization to improve the accuracy of 6-DoF VO. On the other hand, relative motion information from odometry can also be used to improve the global pose regression accuracy.
In summary, our main contributions are as follows:

\begin{enumerate}[(1)]
\item We demonstrate the architecture consisting of the CNN-based feature extraction sub-networks (CNN1), the RCNNs-type relative and global pose regression sub-networks (named RCNN1 and RCNN2 respectively), and finally Fully-connected fusion layers (FCFL) fuse the global and relative poses by connecting these sub-networks to each other.
\item The training strategy: we firstly train the feature extraction and relative pose estimation sub-networks from a sequence of raw RGB images, and then the whole architecture is trained in an end-to-end manner to fill the rest of the pose regression sub-networks according to different scenes.
\item We design two loss functions to improve the accuracy of our networks. For training the relative sub-networks, the CTC is employed to enforce the temporal geometric consistency between each other within a batch of frames. For training the whole networks, we minimize both CTC and the pose MSE.
\item We evaluate our networks using 7-Scenes and KITTI datasets, and the results show it achieves state-of-the-art performance for learning-based monocular camera localization.
\end{enumerate}

\begin{figure*}[t]
\centering
\includegraphics[scale=0.35]{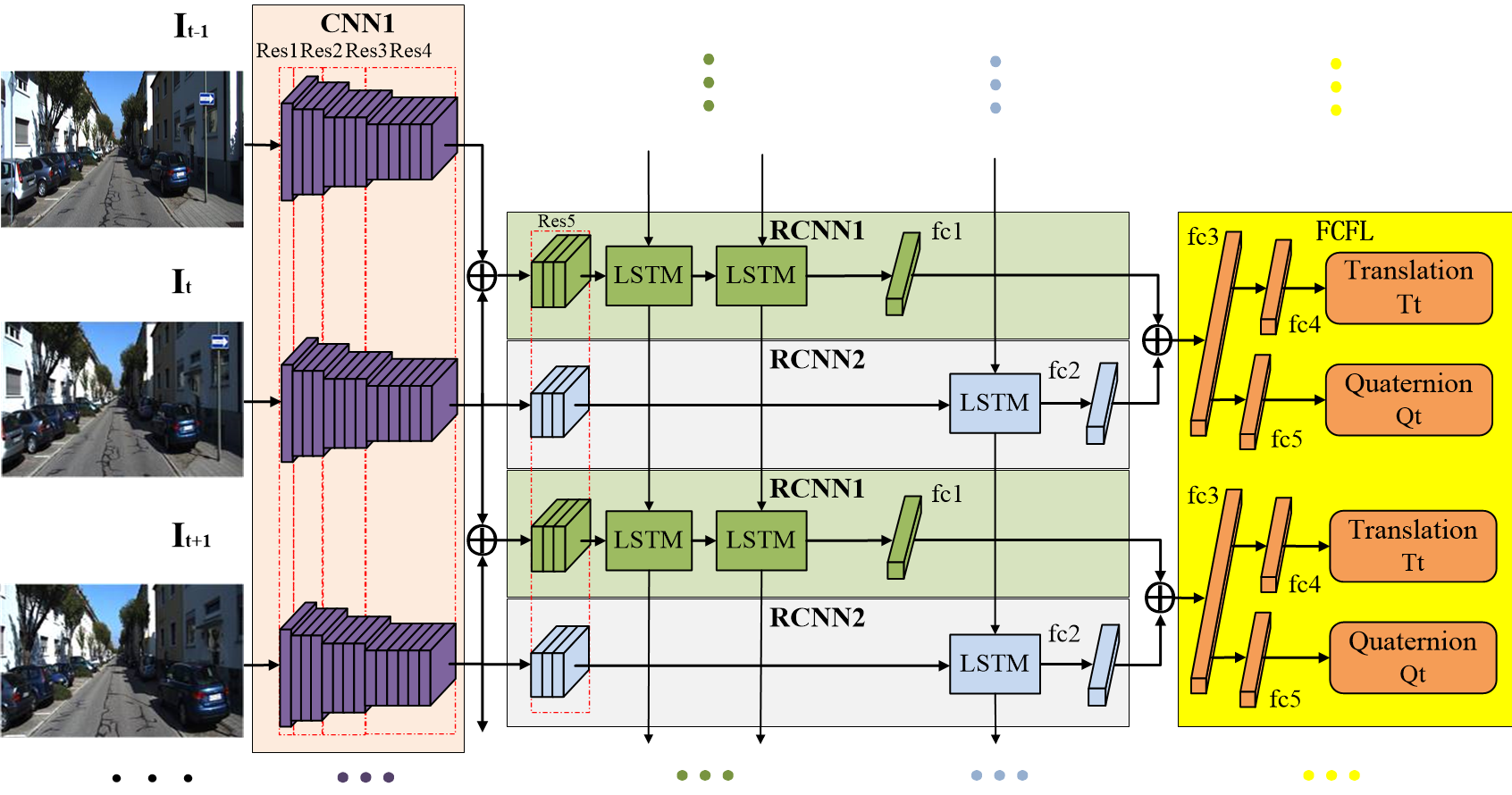}
\caption{Architecture of the proposed learning-based monocular VO system. CNN1 determines the most discriminative feature as an input for the next two RCNNs; RCNN1 estimates the egomotion of the camera and constrict the motion space while regressing the global localization; RCNN2 is competent to model the 3D structural constraints of the environment while learning from the first two assistant networks; Fully-connected fusion layers (FCFL) fuse relative and global networks to improve the VO accuracy.}
\label{fig:fig2}
\end{figure*}

\section{Related Work}\label{sec:review}

Over the past years, there are numerous approaches that have been proposed for visual localization. In this section, we discuss traditional geometry-based and recent learning-based localization approaches.

\subsection{Geometry-based Localization}\label{sec:review:slam}

Geometry-based localization estimates the camera motion among a set of consecutive frames with geometric methods. A variety of geometric methods can be classified into feature-based and direct methods.

\emph{Feature-based methods:} most feature-based methods work by detecting feature points and matching them between consecutive frames. To improve pose accuracy, they minimize the projective geometry errors between 3D feature points of the scene and their projection on the image plane, e.g., PTAM~\cite{4PTAM} is a classical vSLAM system. However, it may suffer from drift since it does not address the principle of loop closing. More recently, the ORB-SLAM algorithm by Mur-Artal et al.~\cite{3ORBSLAM} is state-of-the-art vSLAM system designed for sparse feature tracking and reached impressive robustness and accuracy. In practice, it also suffers from a number of problems such as the inconsistency in initialization, and the drift caused by pure rotation.

\emph{Direct methods:} in contrast, direct methods estimate the camera motion by minimizing the photometric error over all pixels across consecutive images. Engel at al.~\cite{5engel2014lsd}developed LSD-SLAM, which is one of the most successful direct approaches. Direct methods do not provide better tolerance towards changing lighting conditions and often require more computational costs than feature-based methods since they work a global minimization using all the pixels in the image.

\subsection{Learning-based Localization}\label{sec:review:seg}

Even though Deep Neural Networks (DNNs) are not a novel concept, their popularity has grown in recent years due to a great breakthrough that has been achieved in the computer vision community. Inspired by these achievements, lots of learning-based visual relocalization and odometry systems have been widely proposed to improve the 6-DoF pose estimation.

\emph{Visual relocalization:} Learning-based relocalization systems are designed to learn from recognition to relocalization with very large scale classification datasets. For example, Kendall et al. proposed PoseNet~\cite{7kendall2015posenet}, which was the first successful end-to-end pre-trained deep CNNs approach for 6-DoF pose regression. In addition, Clark et al. ~\cite{8walch2017image} introduced deep CNNs with Long-Short Term Memory (LSTM) units to avoid overfitting to training data while PoseNet needs to deal with this problem with careful dropout strategies.

\emph{Visual odometry:} learning-based visual odometry systems are employed to learn the incremental change in position from images. LS-VO~\cite{9costante2018ls} is a CNNs architecture proposed to learn the latent space representation of the input Optical Flow field with the motion estimate task. SfM-Net~\cite{10vijayanarasimhan2017sfm} is a self-supervised geometry-aware CNNs for motion estimation in videos that decomposes frame-to-frame pixel motion in terms of scene and object depth, camera motion and 3D object rotations and translations. Recently, most state-of-the-art deep approaches to visual odometry employ not only CNNs, but also sequence-models, such as long-short term memory (LSTM) units~\cite{11iyer2018geometric}, to capture long term dependencies in camera motion.

More recently, learning-based global and relative networks are designed for 6-DoF global pose regression and odometry estimation from consecutive monocular images. Clark et al.~\cite{14clark2017vidloc} have presented a CNNs+Bi-LSTMs approach for 6-DoF video-clip relocalization that exploits the temporal smoothness of the video stream to improve the localization accuracy of the global pose estimation. Brahmbhatt et al.~\cite{13brahmbhatt2018geometry} proposed a MapNet that enforces geometric constraints between relative poses and absolute poses in network training. Our work is extended to VLocNet~\cite{12valada2018deep}, which incorporated a global and a relative sub-networks. More precisely, even though it has the joint loss function designed for global and relative sub-networks, it is just used to improve the global predictions. Conversely, the global regression results are unable to totally benefit relative networks since its unshared weights are optimized independently without consider the global constraints. Moreover, it considers only a single image as global networks input, which greatly impedes the ability of CNNs to achieve accurate poses. In contrast, we fuse these two streams from both global and relative RCNNS-type sub-networks with joint optimization to benefit the pose prediction.


\section{Proposed Model}\label{sec:algo}

In this section, we detail our learning-based global and relative fusion framework for jointly estimating global pose and odometry from consecutive monocular images. The proposed networks are shown in Fig.~\ref{fig:fig2}.

\subsection{Network Architecture}

\subsubsection{CNN-based feature extraction networks (CNN1)}

In order to learn effective features that are suitable for the global and relative pose estimation problem automatically, CNN-based feature extraction networks are developed to perform feature extraction on the monocular RGB image. We build upon this networks using the first four residual blocks of the ResNet-50 (named from Res 1 to Res 4) ~\cite{15he2016deep}. Each residual unit has a bottleneck architecture consisting of 1$\times$1 convolution, 3$\times$3 convolution, 1$\times$1 convolution layers. Each of the convolutions is followed by batch normalization, scale and Exponential Linear Units (ELUs)~\cite{16clevert2015fast}.

\subsubsection{RCNNs-type relative sub-networks (RCNN1)}

Following the feature extraction networks, the deep RCNNs are designed to model dynamics and relations among a sequence of CNNs features. It takes CNNs features from a consecutive monocular RGB images as input, and then the concatenate features from them are fed into the last residual blocks of the ResNet-50 (Res 5). Note that the output dimension of this layer is W$\times$H$\times$1024. As described in DeepVO~\cite{17wang2018end}, two Long Short-Term Memory (LSTMs)~\cite{18zaremba2014learning} are employed as RNNs to find and exploit correlations among images taken in long trajectories and each of the LSTM layers has 1000 hidden states. The RCNNs output pose estimation at each time step with a fully-connected layer fc1 whose dimension is 1024.

\subsubsection{RCNNs-type global sub-networks (RCNN2)}

We also feed the previous CNNs features to the last residual blocks of the ResNet-50 (Res 5) and reshape LSTM’s output to a fully-connected layer fc2, whose dimension is 1024. It corresponds in shape to the output of the relative RCNNs unit before the fusion stage. Note that the cell of LSTM stores the past few global poses and therefore it is able to improve the predicted pose accuracy of current image.

\subsubsection{Fully-connected fusion layers (FCFL)}

Finally, the following fusion stage concatenates features from the two relative and global sub-networks, and reshapes its output to 1024, namely fc3. We also add two inner-product layers for regressing the translation ${T_k}$ and quaternion ${Q_k}$, namely fc4 and fc5. Obviously, the dimensions of fc4 and fc5 layers are 3 and 4, respectively.

\subsection{Temporal Geometric Consistency Loss}

Here, we introduce CTC that are based on the fundamental concepts of composition of rigid-body transformations. Fig.~\ref{fig:fig4} shows a sequential set of frames $F = \left( {{I_0},{I_1},{I_2},{I_3},{I_4}} \right)$, where we note that temporal length K=5. Note that ${P_i} = ({Q_i},{T_i})$ is a 6-DoF predicted pose, where ${T_i}$ and ${Q_i}$ denote the translation and quaternion of frame $i$, respectively.  We train the networks to predict the transforms between each other: $\left[ {{P_{01}},{P_{12}},{P_{23}},{P_{34}},{P_{02}},{P_{24}},{P_{04}}} \right]$. As an example, the predicted transform ${P_{01}}$ from ${I_{0}}$ to ${I_{1}}$ should be equal to the product of the two ${\mathord{\buildrel{\lower3pt\hbox{$\scriptscriptstyle\frown$}}
\over P} _0}$ and ${\mathord{\buildrel{\lower3pt\hbox{$\scriptscriptstyle\frown$}}
\over P} _1}$ transforms, where ${\mathord{\buildrel{\lower3pt\hbox{$\scriptscriptstyle\frown$}}
\over P} _i}$ indicates the ground truth of frame $i$, thus:

\begin{figure}[t]
\centering
\includegraphics[scale=0.25]{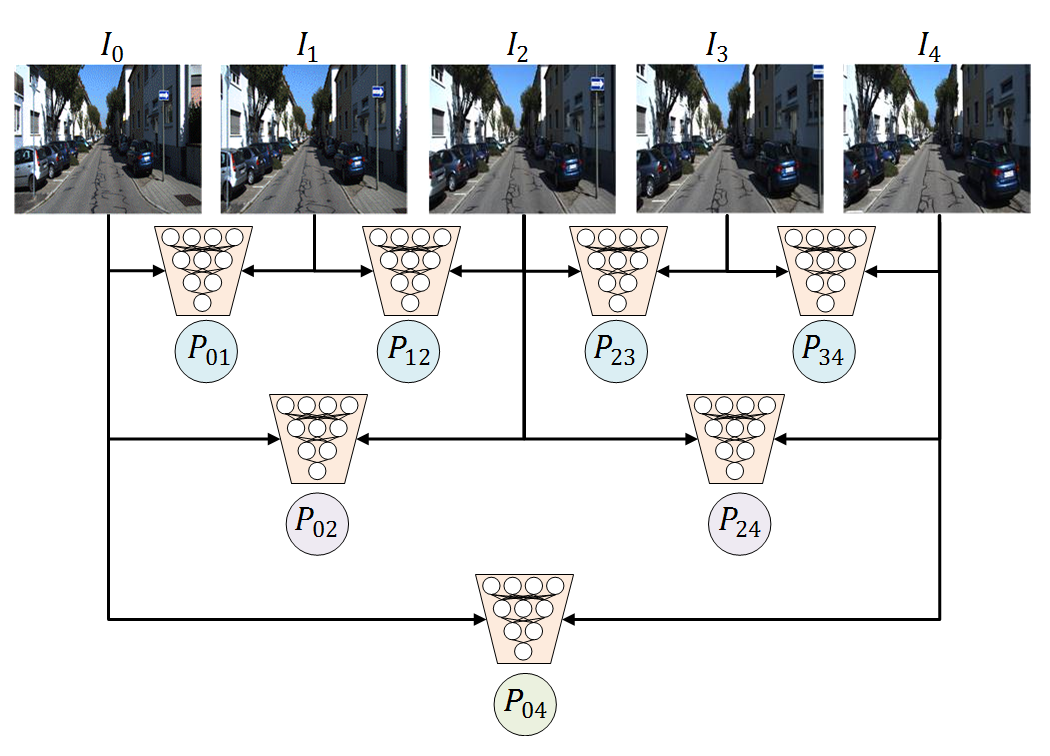}
\caption{Architecture of CTC. It represents temporal geometric consistency of the consecutive frames.}
\label{fig:fig4}
\end{figure}

\begin{equation}\label{eq:1}
{P_{01}} = {\mathord{\buildrel{\lower3pt\hbox{$\scriptscriptstyle\frown$}}
\over P} _1}{\mathord{\buildrel{\lower3pt\hbox{$\scriptscriptstyle\frown$}}
\over P} _0}^{ - 1} = {\mathord{\buildrel{\lower3pt\hbox{$\scriptscriptstyle\frown$}}
\over P} _{01}}
\end{equation}

Note that using Eq.(1) in practice, there exist errors in the predicted and ground truth, so we have CTC functions:

\begin{equation}\label{eq:2}
\begin{array}{l}
{L_0} = \left\| {{P_{01}} - {{\mathord{\buildrel{\lower3pt\hbox{$\scriptscriptstyle\frown$}}
\over P} }_{01}}} \right\|_2^2,{L_1} = \left\| {{P_{12}} - {{\mathord{\buildrel{\lower3pt\hbox{$\scriptscriptstyle\frown$}}
\over P} }_{12}}} \right\|_2^2\\
{L_2} = \left\| {{P_{23}} - {{\mathord{\buildrel{\lower3pt\hbox{$\scriptscriptstyle\frown$}}
\over P} }_{23}}} \right\|_2^2,{L_3} = \left\| {{P_{34}} - {{\mathord{\buildrel{\lower3pt\hbox{$\scriptscriptstyle\frown$}}
\over P} }_{34}}} \right\|_2^2\\
{L_4} = \left\| {{P_{02}} - {{\mathord{\buildrel{\lower3pt\hbox{$\scriptscriptstyle\frown$}}
\over P} }_{02}}} \right\|_2^2,{L_5} = \left\| {{P_{24}} - {{\mathord{\buildrel{\lower3pt\hbox{$\scriptscriptstyle\frown$}}
\over P} }_{24}}} \right\|_2^2\\
{L_6} = \left\| {{P_{04}} - {{\mathord{\buildrel{\lower3pt\hbox{$\scriptscriptstyle\frown$}}
\over P} }_{04}}} \right\|_2^2
\end{array}
\end{equation}

where $\left\|  \cdot  \right\|_2^2$ is MSE.	So the relative loss function which consists of Eq.(2) are shown as:

\begin{equation}\label{eq:3}
\theta  = \mathop {\arg \min }\limits_\theta  \frac{1}{N}\sum\limits_{i = 1}^N {\sum\limits_{k = 0}^6 {(L_k^i)} }
\end{equation}

where $\theta $ is the relative or global RCNNs parameters and N is the number of samples. We use this optimization Eq.(3) to train our RCNNs sub-networks. Note that, these constrains can be equal to a Local Bundle Adjustment in traditional vSLAM system~\cite{3ORBSLAM}, also known as windowed optimization. It is an efficient way to maintain a good quality pose over a local number of frames. So the CTC here are better strategies to learn about spatial relations of the environment.
To train our 6-DoF end-to-end pose regression system, we can jointly use the global and relative loss function as follows:

\begin{equation}\label{eq:4}
w = \mathop {\arg \min }\limits_w \frac{1}{N}\sum\limits_{i = 1}^N {\left\{ {\sum\limits_{k = 0}^6 {(L_k^i) + \sum\limits_{j = 0}^4 {\left\| {{P^i}_j - {{\mathord{\buildrel{\lower3pt\hbox{$\scriptscriptstyle\frown$}}
\over P} }^i}_j} \right\|_2^2} } } \right\}}
\end{equation}

where $\omega$ is the networks parameters. It is obvious that Eq.(4) tries to minimize the Euclidean distance between the ground truth pose and estimated one while enforcing the geometric consistency between each other within a batch of frames.

\subsection{Training Strategy}

We firstly initialize the CNN1 and RCNN1 from a sequence of raw RGB images using the optimization Eq.(3). In particular, RCNN1 directly replace fc3 with fc1 and estimate the pose from fc4 and fc5 for the time being.

\begin{figure}[t]
\centering
\includegraphics[scale=0.25]{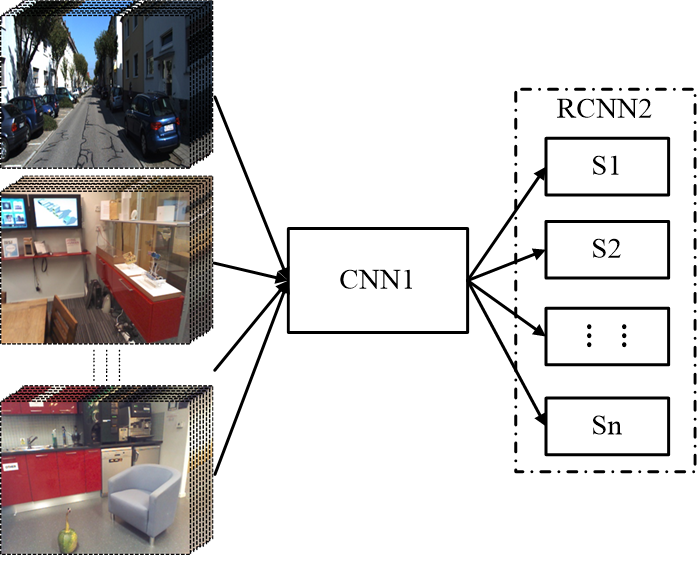}
\caption{Illustration of training for the global networks from different scenes. CNN1 determines the most discriminative feature and RCNN2 learns from  different scenes for saving their landmark Si.}
\label{fig:fig3}
\end{figure}

For initializing RCNN2, we observe that the most global pose regression~\cite{7kendall2015posenet} can only be determined in a known training environment. So it is time consuming to retrain the whole networks according to different scenes. As shown in Fig.~\ref{fig:fig3}, in order to retrain our deep model faster, different scenes are fed into the common CNN1 to produce an effective feature in the monocular image, which is then passed through individual RCNN2 to learn for saving their landmark Si. Thereby, we only need to retrain the RCNN2 for different scenes and each image still yields an accurate pose estimate at each Si through the networks. Note that, RCNN2 also directly replaces fc3 with fc2 and regress the pose using the optimization Eq.(4).

Up to now, we achieve pretrained weights for CNN1, RCNN1 and RCNN2. Finally, the whole architecture is trained and refined in an end-to-end manner via the optimization Eq.(4).

\section{Experimental Evaluation}

In this section, we evaluate our proposed networks in comparison to the state-of-the-art algorithms on both indoor and outdoor datasets, followed by detailed analysis on the architectural decisions and finally, we demonstrate the best temporal length.

\subsection{ Evaluation Datasets}

We evaluate our networks on two well-known datasets: Microsoft 7-Scenes~\cite{19shotton2013scene} and KITTI Visual Odometry benchmark~\cite{20geiger2012we}. We follow the original train and test splits provided by other literatures to facilitate comparison and benchmarking.

\subsubsection{Microsoft 7-Scenes} It is a dataset that collect RGB-D images from seven different scenes in an indoor office environment. All scenes were recorded from a handheld Kinect RGB-D camera at 640$\times$480 resolution. The dataset provides the ground truth poses extracted using KinectFusion. Each sequence was recorded with motion blur, perceptual aliasing and textureless features in the room, thereby making it a challenging dataset for relocalization and tracking.

\subsubsection{KITTI Visual Odometry benchmark} It consists of 22 stereo sequences and they provide 11 sequences (00-10) with ground truth trajectories for training and 11 sequences (11-21) without ground truth for evaluation. This high-quality dataset was recorded with long sequences of varying speed, including a set of 41000 frames captured at 10 \emph{fps} and a total driving distance of 39.2 \emph{km} with frequent loop closures which are of interest in SLAM. So it is very popular for the monocular Visual Odometry algorithms.

\subsection{Network Training}

The network models were implemented with the TensorFlow framework and trained with NVIDIA GTX 1080 GPUs and Intel Core i7 2.7GHz CPU. Adam optimizer was employed to train the networks for up to 2000 epochs with parameter ${\beta _1}$ = 0.9 and ${\beta _2}$ = 0.999. The learning rate started from 0.001 and decreased by half for every 1/5 of total iterations. The temporal length K fed to the relative and global pose estimator is 5. The size of image used by the networks is 224$\times$224 pixel. Thus, our per-frame runtime for each pose inference is between 45 ms and 65 ms.

\subsection{Microsoft 7-Scenes Datasets}

In this experiment, we compare the performance of our networks with other state-of-the-art deep learning-based relocalization and tracking methods, namely PoseNet~\cite{7kendall2015posenet}, DeepVO~\cite{21mohanty2016deepvo} and VLocNet~\cite{12valada2018deep}. In order to implement fair qualitative and quantitative comparison, we use the same 7-Scenes datasets  for training and testing as described in ~\cite{12valada2018deep}. For each scene, we show their median translational ${t_{rel}}$: ($m$) and rotational ${r_{rel}}$: ($^\circ $) errors in Table~\ref{tab:comp1}, respectively.

\renewcommand\arraystretch{1.2}
\begin{table}
\centering
\begin{threeparttable}
  \caption{MEDIAN ERRORS ON MICROSOFT 7-SCENES}
  \label{tab:comp1}
  \centering
  \begin{tabular}{p{1.2cm}p{0.4cm}p{0.4cm}p{0.4cm}p{0.4cm}p{0.4cm}p{0.4cm}p{0.4cm}p{0.4cm}}
    \hline
    \multirow{2}{*}{\textbf{Scene}} & \multicolumn{2}{c}{\textbf{PoseNet}} & \multicolumn{2}{c}{\textbf{DeepVO}} &
    \multicolumn{2}{c}{\textbf{VLocNet}} & \multicolumn{2}{c}{\textbf{Ours}}\\
    \cline{2-9}
     & \multicolumn{1}{p{0.25cm}}{${t_{rel}}$} & \multicolumn{1}{p{0.25cm}}{${r_{rel}}$} & \multicolumn{1}{p{0.25cm}}{${t_{rel}}$} & \multicolumn{1}{p{0.25cm}}{${r_{rel}}$} & \multicolumn{1}{c}{${t_{rel}}$} & \multicolumn{1}{p{0.3cm}}{${r_{rel}}$} & \multicolumn{1}{c}{${t_{rel}}$} & \multicolumn{1}{p{0.3cm}}{${r_{rel}}$}\\
    \hline
            \small{Chess} & 0.32 & 8.12 & 0.06 & 2.61 & 0.036 & \textbf{1.70} & \textbf{0.016} & 1.72\\
            \small{Fire} & 0.47 & 14.4 & 0.10 & 4.33 & 0.039 & 5.33 & \textbf{0.011} & \textbf{2.19}\\
            \small{Heads} & 0.29 & 12.0 & 0.35 & 7.11 & 0.046 & 6.64 & \textbf{0.017} & \textbf{3.56}\\
            \small{Office} & 0.48 & 7.68 & 0.10 & 3.11 & 0.039 & \textbf{1.95} & \textbf{0.024} & \textbf{1.95}\\
            \small{Pumpkin} & 0.47 & 8.42 & 0.11 & 3.30 & 0.037 & 2.28 & \textbf{0.022} & \textbf{2.27}\\
            \small{RedKitchen} & 0.59 & 8.64 & 0.10 & 2.58 & 0.039 & 2.20 & \textbf{0.018} & \textbf{1.86}\\
            \small{Stairs} & 0.47 & 13.8 & 0.45 & 9.18 & 0.097 & 6.47 & \textbf{0.017} & \textbf{4.79}\\
    \hline
    \footnotesize{Average} & 0.44 & 10.4 & 0.18 & 4.60 & 0.048 & 3.80 & \textbf{0.018} & \textbf{2.62}\\
  \hline
\end{tabular}
\end{threeparttable}
\end{table}

It shows that our networks outperform previous CNN-based PoseNet by 95.9$\%$ in positional error and 74.8$\%$ in orientation error. Taking Pumpkin as an example, we achieve a positional error reduction from 0.47m for PoseNet to 0.022m for our method. The reason is that PoseNet always results in noisy predictions on single image. In contrast, the RCNNs in our networks constrict the motion space while using sequential images to improve global relocalization accuracy. Therefore, this experiment results validate that our networks have the effectiveness of using geometric constraints from consecutive images for improving relocalization accuracy. Furthermore, it can be seen that the proposed networks significantly outperform the DeepVO approach in all of the test scenes, resulting in a 90$\%$ and 43$\%$ boost in position and orientation accuracy, respectively. The DeepVO network tries to regress the VO but probably suffers from high drifts. The reason is that the orientation changes in the training data are usually small and orientation is more prone to overfitting. However, our system reduces the drift over time due to the global pose regression strategy as done in the traditional visual SLAM system. In addition, our networks also perform better than VLocNet, and the orientation and positional errors are reduced by more than 31$\%$ and 62$\%$, respectively. The main reason we find from VLocNet is that their global pose regression and visual odometry networks are predicted independently. But in our framework, we do fuse the results from global regression and relative pose estimation. In summary, these experimental results validate that our strategy is able to filter out the noises by fusing a series of measurements observed from global and relative networks over time.

\subsection{KITTI Datasets}

\renewcommand\arraystretch{1.2}
\begin{table}
\centering
\begin{threeparttable}
  \caption{RESULTS ON KITTI SEQUENCES}
  \label{tab:comp2}
  \centering
  \begin{tabular}{ccccccc}
    \hline
    \multirow{2}{*}{\textbf{Seq.}} & \multicolumn{2}{c}{\textbf{DeepVO}} & \multicolumn{2}{c}{\textbf{L-VO3}} &
    \multicolumn{2}{c}{\textbf{Ours}}\\
    \cline{2-7}
     & \multicolumn{1}{c}{${t_{rel}}$} & \multicolumn{1}{c}{${r_{rel}}$} & \multicolumn{1}{c}{${t_{rel}}$} & \multicolumn{1}{c}{${r_{rel}}$} & \multicolumn{1}{c}{${t_{rel}}$} & \multicolumn{1}{c}{${r_{rel}}$} \\
    \hline
            03 & 6.72 & 6.46 & 3.18 & \textbf{1.31} & \textbf{1.93} & 1.95\\
            04 & 6.33 & 6.08 & 2.04 & 0.81 & \textbf{0.10} & \textbf{0.25}\\
            05 & 3.35 & 4.93 & \textbf{2.59} &0.99 & 2.51 & \textbf{0.91}\\
            06 & 7.24 & 7.29 & 1.38 & \textbf{0.95} & \textbf{0.30} & 0.99\\
            07 & 3.52 & 5.02 & 2.81 & 2.54 & \textbf{1.53} & \textbf{2.11}\\
            10 & 9.77 & 10.2 & 4.38 & 3.12 & \textbf{3.63} & \textbf{3.00}\\
    \hline
    Average & 6.15 & 6.66 & 2.73 & 1.62 & \textbf{1.67} & \textbf{1.54}\\
  \hline
\end{tabular}
\end{threeparttable}
\end{table}

Next, we additionally deploy experiments in an outdoor environment for analyzing the large-scale VO performance. KITTI is much larger than typical indoor datasets like 7-Scenes, where sequence 00, 02, 08 and 09 are used for training the RCNNs-type relative sub-networks. As described in~\cite{17wang2018end}, the trajectories are segmented to different lengths to generate almost 7410 samples in total for training. The trained models are tested on the sequence 03, 04, 05, 06, 07 and 10. As shown in Table~\ref{tab:comp2}, the performance of the our networks is analyzed according to the KITTI VO/SLAM evaluation metrics, where ${t_{rel}}$: ($\%$) and ${r_{rel}}$: ($^\circ /$100m) are averaged Root Mean Square Errors (RMSE) of the translational and rotational drifts for all subsequences of lengths ranging from 100 to 800 meters with different speeds.

\begin{figure*}[t]
\centering
\includegraphics[scale=0.18]{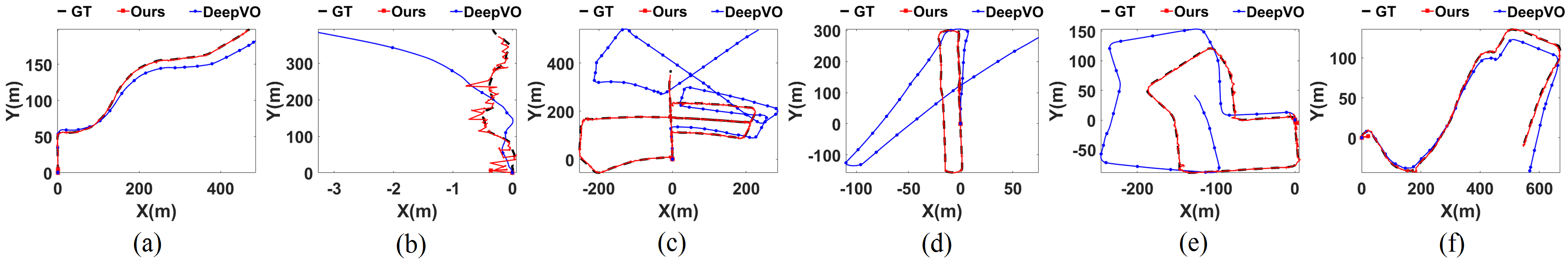}
\caption{Trajectories of results on the testing Sequence (a) 03, (b) 04, (c) 05, (d) 06, (e) 07 and (f) 10 of the KITTI VO benchmark.}
\label{fig:fig5}
\end{figure*}

 Table~\ref{tab:comp2} shows quantitative comparison against two state-of-the-art VO approaches including L-VO3~\cite{22zhao2018learning} and RCNNs-type DeepVO~\cite{17wang2018end}. The proposed method significantly outperforms the DeepVO approach in all of the test sequences, resulting in a 71$\%$ and 76$\%$ boost in translation and rotation accuracy, respectively. As shown in Fig.~\ref{fig:fig5}, DeepVO suffers from high drifts as the length of the trajectory increases and the errors of the rotation significantly increase because of significant changes on rotation during car driving. Unlike that, our networks produce relatively accurate and consistent trajectories against to the ground truth. These owe to the global and relative architecture with the proposed CTC loss. In addition, it is able to overcome the performance of state-of-the-art learning-based L-VO3. Although some errors are slightly worse than that of the L-VO3, this may be due to the fact that our networks are trained without enough data to cover the velocity and orientation variation. Finally, we can see that the absolute scale to each sequence is completely maintained during the end-to-end training.


\subsection{Ablation Studies}
In this section, we present additional ablation studies on performances with respect to considering various architectural components and temporal length K.


\begin{figure}[t]
\centering
\includegraphics[scale=0.2]{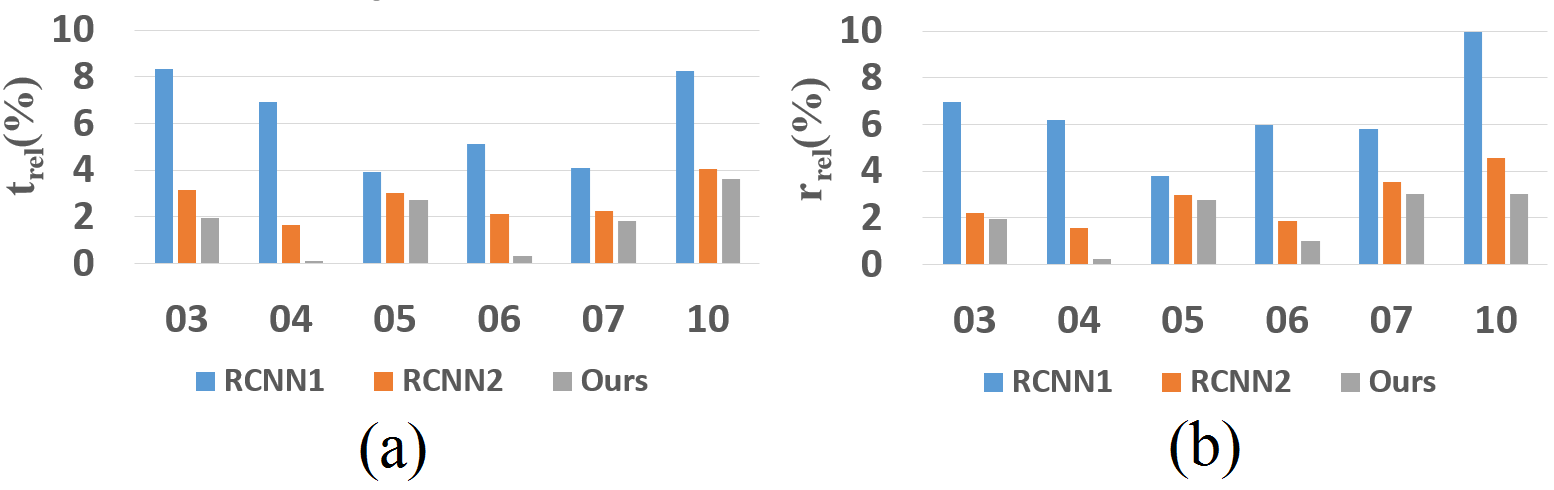}
\caption{Comparative analyses of average (a) translation and (b) rotation RMSE using various architectures on KITTI sequences.}
\label{fig:fig6}
\end{figure}

In order to validate the effectiveness of our joint architecture, we compare our networks against relative-only (RCNN1) and global-only (RCNN2) architectures. The quantitative rules can be found in Section 4.4. In particular, RCNN1 directly replaces fc3 with fc1 and estimates the pose from consecutive images. While RCNN2 also directly replaces fc3 with fc2 and regresses the pose. They are trained using the loss function Eq.(3) and Eq.(4) respectively, and temporal length K equals to 5 as well. It is observed that compared with the RCNN1, the pose generated from RCNN2 is more accurate. A possible explanation is that the global networks reduce the serious drift since it has the ability to relocalization with previous observation for the long-term prediction. While the relative networks only focus on motion from 2D or 3D optical flow, which is hard to efficiently model 3D structural constraints with limited training samples in complex environments. Compared to the RCNN1 and RCNN2 as shown in Fig.~\ref{fig:fig6}, our approach predicts more precise pose. This is more evident if we fuse streams from both global and relative sub-networks to benefit the long-term pose prediction.

\begin{figure}[t]
\centering
\includegraphics[scale=0.25]{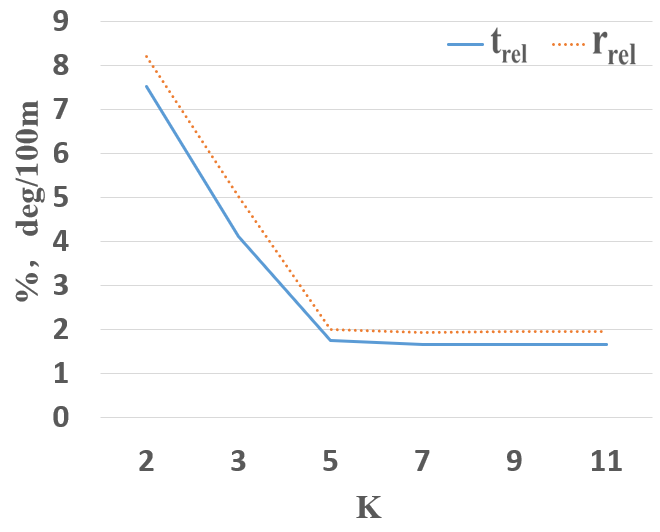}
\caption{Comparation of average translation ${t_{rel}}$ and rotation ${r_{rel}}$ RMSE with respect to various temporal length K.}
\label{fig:fig7}
\end{figure}

In addition, we provide a performance comparison with respect to various temporal lengths. Fig.~\ref{fig:fig7} shows our networks with different K values changed from 2 to 11 and their corresponding average translation and rotation RMSE of the six KITTI sequences. We observe that the localization errors descend as the length of the sequential frames increases. However, the accuracy seems to be stable when K is larger than 5. This phenomenon is due to the fact that the covisible constraint between 1-th frame and K-th frame become weak when K is large enough. Furthermore, we find that training such networks, especially when K is larger than 5, requires more training data to generalize well in unseen data and avoid overfitting. Therefore, we conclude that the temporal length k=5 is the best configuration for VO task.

\section{Conclusion}\label{sec:conc}

In this paper, we addressed the challenge of learning-based visual localization of a camera or an autonomous system with the novel networks. It mainly consists of CNN-based feature extraction sub-networks that determine the most discriminative feature as an input for the next two RCNNs, RCNNs-type relative sub-networks that estimate the egomotion of the camera and constrict the motion space while regressing the global localization, and RCNNs-type global sub-networks that are competent to model the 3D structural constraints of the environment while learning from the first two assistant networks. Finally, it fuses and jointly optimizes the relative and global networks to improve VO accuracy. Furthermore, we employ the CTC loss function for training the relative and global RCNNs. The indoor and outdoor experimental evaluations indicate that our networks can produce accurate localization and be adopted to maintain a large feature map for drift correction under long range pose estimation. In the next step, we plan to extend the ability of global networks to work under any unknown environment and promote the robustness of place recognition in cases where illumination and appearance change dramatically.

\bibliographystyle{named}
\bibliography{IEEEexample}

\end{document}